\documentclass{anstrans}
\title{Native Fortran Implementation of TensorFlow-Trained Deep and Bayesian Neural Networks}
\author{Aidan Furlong,$^{*}$ Xingang Zhao,$^{\dagger}$ Robert Salko,$^{\ddagger}$ Xu Wu$^{*}$}

\institute{
$^{*}$Department of Nuclear Engineering, North Carolina State University, Burlington Engineering Laboratories, 2500 Stinson Dr., Raleigh, NC 27695, ajfurlon@ncsu.edu, xwu27@ncsu.edu
\and
$^{\dagger}$Department of Nuclear Engineering, University of Tennessee, Knoxville, Zeanah Engineering Complex, 863 Neyland Dr., Knoxville, TN 37996, xzhao47@utk.edu
\and
$^{\ddagger}$Nuclear Energy and Fuel Cycle Division, Oak Ridge National Laboratory Oak Ridge, TN 37830, salkork@ornl.gov
}

\usepackage{graphicx} % allows inclusion of graphics
\usepackage{booktabs} % nice rules (thick lines) for tables
\usepackage{microtype} % improves typography for PDF
\usepackage{siunitx}

\newcommand\blfootnote[1]{%
  \begingroup
  \begin{NoHyper}
  \renewcommand\thefootnote{}\footnote{#1}%
  \addtocounter{footnote}{-1}%
  \end{NoHyper}
  \endgroup
}

\begin{document}
%%%%%%%%%%%%%%%%%%%%%%%%%%%%%%%%%%%%%%%%%%%%%%%%%%%%%%%%%%%%%%%%%%%%%%%%%%%%%%%%

\blfootnote{
Notice: This manuscript has been authored by UT-Battelle LLC under contract DE-AC05-00OR22725 with the US 
Department of Energy (DOE). The US government retains and the publisher, by accepting the article for 
publication, acknowledges that the US government retains a nonexclusive, paid-up, irrevocable, worldwide 
license to publish or reproduce the published form of this manuscript, or allow others to do so, for US 
government purposes. DOE will provide public access to these results of federally sponsored research in 
accordance with the DOE Public Access Plan (https://www.energy.gov/doe-public-access-plan).
}

\section{Introduction}

Over the past decade, the investigation of machine learning (ML) within the field of nuclear engineering has grown significantly. Nuclear engineering applications for ML range from the prediction of fuel lattice parameters to anomaly detection in power plant equipment. Most of the previous ML-related work investigating these solutions has been performed on a research scale with carefully designed ML experiments. Many of the methods developed from this ML research are approaching maturity~\cite{qi2025machine}, so studies will shift toward practical application. This next phase of investigation will help determine the feasibility and usefulness of ML model implementation in a production setting.

Before these ML models are implemented within a nuclear code, a basic assessment of direct compatibility is first required. Several of the codes used for reactor design and assessment---including MCNP~\cite{TechReport_2023_LANL_LA-UR-22-33103RisingArmstrongEtAl}, CASMO/SIMULATE~\cite{rhodes2006casmo}, RELAP5~\cite{fletcher1992relap5}, and CTF~\cite{salko2020ctf}, among others---are primarily written in the Fortran language. Conversely, most ML model design, training, testing, and manipulation is performed in Python using either TensorFlow or PyTorch. The trained models are saved in a user-defined file format, such as HDF5 or ONNX, to allow for rapid loading in other ML frameworks. A custom framework is necessary to read these files and effectively ``load'' a model's architecture within Fortran. 

This work presents a public framework that enables the direct implementation of TensorFlow-trained models within Fortran using the HDF5 file format to store the model. This framework provides a fully native solution for deep neural networks (DNNs)~\cite{furlong2024fortrandeep} and Bayesian neural networks (BNNs)~\cite{furlong2024fortranbayesian} within Fortran without the dependence on Python runtime environments, TensorFlow’s C API, or ONNX conversion. This simple and lightweight implementation is suitable for applications that require a large volume of predictions (e.g., within iterative solvers). The computational efficiency aspect also provides the opportunity to effectively implement DNN ensembles, which support uncertainty quantification (UQ) in addition to the BNN outputs' inherent UQ capabilities. This allows for confidence information to be provided for every prediction, better informing on their qualities. Although this work was originally developed to be used in CTF, it is completely code agnostic and can be used in any Fortran deployment.

In this summary, the framework is described in detail, and this description is followed by verification comparing the Fortran predictions with those made in Python with TensorFlow. This verification was accomplished with two demonstration problems: (1) a noisy sinusoid and (2) the prediction of critical heat flux (CHF) values.

%%%%%%%%%%%%%%%%%%%%%%%%%%%%%%%%%%%%%%%%%%%%%%%%%%%%%%%%%%%%%%%%%%%%%%%%%%%%%%%%
\section{Implementation}

This framework is designed for simplicity in its implementation within the intended application. In this spirit, all components related to model loading, data preprocessing, and prediction are located within a single module---\texttt{dnn\_module.f90} in the case of DNNs, or \texttt{bnn\_module.f90} in the case of BNNs. In terms of files, the TensorFlow output HDF5 model file is required. Because the training data was standardized in most cases, the inputs used to create a prediction must also be standardized. The information associated with standardization (x\_mean, x\_std, y\_mean, and y\_std) can easily be placed in a \texttt{metadata.h5} file from the TensorFlow script used to train the model. Making a prediction involves four steps:

\begin{itemize}
    \item Define the model structure (layers, neurons, activations)
    \item Load the weights and biases from the model file
    \item Load the standardization parameters from the metadata~file
    \item Standardize the data and make a prediction
\end{itemize}

In the section of the target script that will call \texttt{dnn\_module.f90} to make predictions, the user defines critical aspects of the model and directs the flow of data through the prediction process. The model initializer is first called to define the model's architecture, \texttt{initialize([$N^{(l)}$, $N^{(l+1)}$, ..., $N^{(L)}$])}, where $N$ denotes the number of neurons for each layer starting at the input layer $l$ and ending at the output layer $L$. The actual weights and biases are then loaded by calling \texttt{load\_weights(model\_path)}. The final step of configuring the model is simply assigning the activation functions. This step is accomplished by associating a procedure pointer within the derived type for activations, as in \texttt{layer\_activation($l$)\%func => relu\_fn}. The \texttt{relu\_fn} is one of five pre-included activation functions, along with \texttt{elu\_fn}, \texttt{selu\_fn}, \texttt{softplus\_fn}, and \texttt{tanh\_fn}. Other activation functions may be added at the discretion of the user.

If standardization is necessary, then the metadata file is loaded by calling \texttt{load\_metadata(metadata\_path)}. Once the metadata file has been loaded, \texttt{standardize(x\_data, x\_mean, x\_std)} may be called for each input value (or vector if the user is making multiple predictions from a single call instead of calling for each prediction). This function returns \texttt{x\_data} after performing the standardization. Finally, the prediction may be obtained by calling \texttt{y\_pred = predict(x\_data)}. Depending on whether the model was trained with a standardized target, calling \texttt{unstandardize(y\_data, y\_mean, y\_std)} may be necessary. Listing \ref{lst:sinusoid_example} is an example of what this workflow could look like.

\begin{lstlisting}[style=fortran, caption={Example code in function form for the sinusoid problem (with inputs \texttt{x1} and \texttt{x2}). In practice, all initialization and model configuration would occur once in the main program to avoid repeating for every function call.}, label={lst:sinusoid_example}]
real function predict_sinusoid_y(x1, x2, x_mean, x_std, y_mean, y_std) result(y_pred)
    real, intent(in) :: x1, x2
    real, intent(in) :: x_mean(2), x_std(2), y_mean(2), y_std(2)
    real :: x_data(2) ! Hardcoded for two inputs
    real :: y_temp
    
    ! Neurons in each layer [Input, Hidden1, ..., Output]
    call initialize_network([2,16,16,1])
    
    ! Read in weights, biases, and scaler info
    call load_weights(model_path)
    call load_metadata(metadata_path, x_mean, y_mean, x_std, y_std)
    
    ! Assign activations, defaults to ReLU
    layer_activations(1)%func => relu_fn
    layer_activations(2)%func => relu_fn
    layer_activations(3)%func => no_activation
    
    ! Combine input data into a single array
    x_data(1) = x1
    x_data(2) = x2
    
    ! Standardize input vectors if needed
    call standardize(x_data(1), x_mean(1), x_std(1))
    call standardize(x_data(2), x_mean(2), x_std(2))
    
    ! Make and collect prediction
    y_temp = predict(x_data)
    
    ! Transform output back to physical dimensions
    call unstandardize(y_temp, y_mean(1), y_std(1))

    ! Final prediction
    y_pred = y_temp
end function predict_sinusoid_y
    
\end{lstlisting}

%%%%%%%%%%%%%%%%%%%%%%%%%%%%%%%%%%%%%%%%%%%%%%%%%%%%%%%%%%%%%%%%%%%%%%%%%%%%%%%%
\section{Verification}

Verification is necessary to ensure that the predictions made by the Fortran implementation accurately match those produced in the Python TensorFlow environment. In this study, verification was conducted using two reasonably complex test cases followed by a nuclear engineering application. The first test case involved a transformed sinusoidal function with random Gaussian noise, taking two inputs and predicting the corresponding \textit{y}-value. Another test case featured a highly nonlinear function with three inputs, again using the \textit{y}-value as the target. The nonlinear regression case is included in the public repository but is omitted here for space considerations. These cases were specifically designed to introduce sufficient complexity, ensuring that agreement between the Fortran and TensorFlow predictions is not merely due to simplicity (e.g., trivially predicting a linear function).

\subsection{Test Case 1: Noisy Sinusoid}

In the case of the noisy sinusoid, a dataset composed of 5{,}000 points was generated with 80\% used for training and 20\% used for testing. Although the goal of these test cases is to assess the agreement between the Fortran and TensorFlow implementations (which could be achieved regardless of the accuracy of the model itself in predicting the test set), the architectures were still optimized prior to training to achieve reasonable error metrics. This was done using a traditional random search space with 1000 prospective configurations. The DNN model used two hidden layers, the BNN model used three hidden layers, and both were trained to 500 epochs.

\subsubsection{Deep Neural Networks}

The results from the DNN model were first considered by comparing 9 relevant error metrics between the implementations, along with the inference timing statistics. Because 20\% of the generated dataset was allocated for testing, predictions were made for a total of 1000 input combinations. The Fortran implementation was able to accomplish this in 5.1\% of the time that the TensorFlow counterpart used. This difference represents a speedup factor of 19.6. A complete listing of the 9 performance metrics is provided in Table \ref{tab:dnn_sinusoid}. Both implementations show good agreement with only small differences. AE denotes the absolute error, APE denotes the absolute percentage error, \textit{rRMSE} denotes the relative root-mean-square error, and $F_{\text{error}}>10\%$ denotes the fraction of relative error values above 10\%.

\begin{table}[ht!]
    \centering
    \caption{Comparison of Fortran-based and TensorFlow-based DNN predictions on the noisy sinusoid problem.}
    \label{tab:dnn_sinusoid}
    \begin{tabular}{lccc}
        \toprule
        \textbf{Metric} & \textbf{Fortran} & \textbf{TensorFlow} & \textbf{Diff.} \\
        \midrule
        Mean AE            & 3.006936  & 3.006934  & 0.000002  \\
        Max AE             & 18.713116 & 18.713146 & $-0.000031$ \\
        Min AE             & 0.002058  & 0.002068  & $-0.000010$ \\
        \midrule
        Mean APE (\%)      & 0.670924  & 0.670924  & 0.000000  \\
        Max APE (\%)       & 3.950175  & 3.950181  & $-0.000006$ \\
        % Min APE (\%)       & 0.000690  & 0.000694  & $-0.000003$ \\
        Std APE (\%)       & 0.614194  & 0.614194  & 0.000000  \\
        \midrule
        \textit{rRMSE} (\%)         & 0.009094  & 0.009094  & 0.000000  \\
        $F_{\text{error}}>10\%$ (\%)   & 0.000000  & 0.000000  & 0.000000  \\
        $R^2$              & 0.998099  & 0.998099  & $-0.000000$ \\
        \midrule
        \textbf{Total predictions} & 1{,}000      & 1{,}000      & -- \\
        \textbf{CPU time (s)} & 0.006927  & 0.136597  & $-0.129670$ \\
        \bottomrule
    \end{tabular}
\end{table}

\begin{figure}[ht!]
    \centering
    \includegraphics[width=\linewidth]{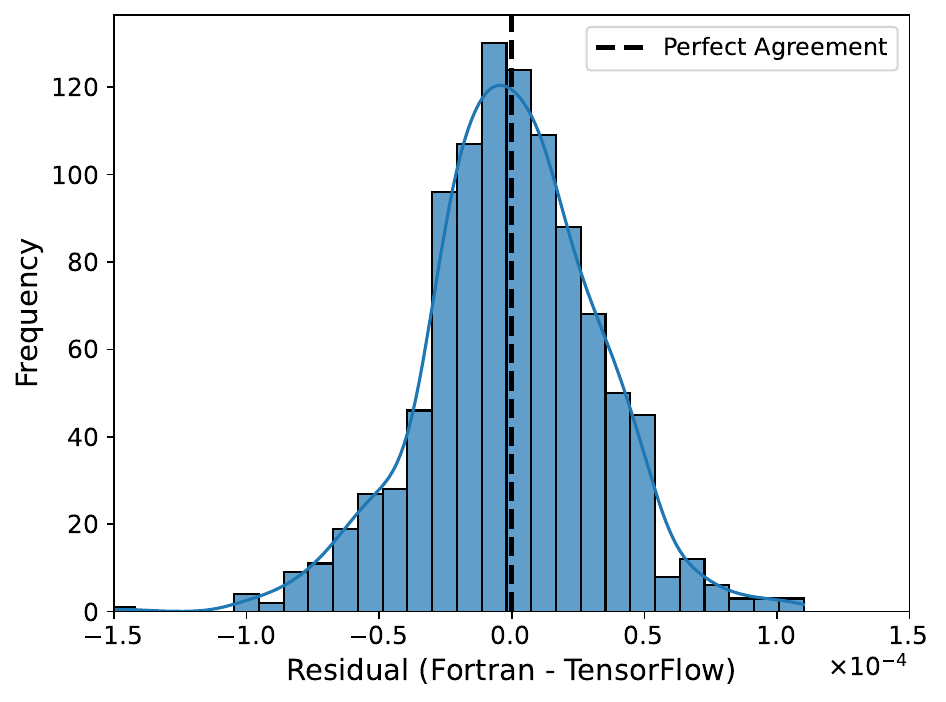}
    \caption{Histogram and KDE plot of the Fortran--Tensorflow DNN residual distribution.}
    \label{fig:dnn_histogram}
\end{figure}

The residuals between the Fortran- and TensorFlow-based predictions were then collected and plotted via histogram and kernel density estimation (KDE), as shown in Fig.~\ref{fig:dnn_histogram}. Both are centered close to zero, suggesting that systematic bias is unlikely; in the absence of systematic bias, residuals are often expected to exhibit a symmetric distribution around zero. The spread is also relatively tight; nearly all residuals are enclosed by $\pm$ $ 1.0 \times 10^{-4}$ on a test set with bounds of 273.6 and 626.3.

\subsubsection{Bayesian Neural Networks}

Comparing the BNN model implementations is less straightforward than comparing the DNN cases because sampling of the posterior distribution is required. This sampling requires randomness, meaning that a reliable random number generator (RNG) is needed. Although RNGs themselves are not conceptually complicated, TensorFlow's default RNG implementation is more complex because of its underlying algorithm and seeding mechanisms. For this reason, obtaining identical values from the custom Fortran and TensorFlow RNGs is not feasible, which can lead to differences between predictions. To mitigate the effect of RNG variability and to ensure a well-approximated posterior distribution, 20,000 samples were taken for each prediction. Totaling 20 million samples, the Fortran implementation took 2.53 min, whereas the TensorFlow counterpart took 20.26 min, meaning that the speedup factor is 8.0. In practice, predictions could be well converged at just 100 samples. In this case, the time elapsed per sample-mean prediction would be 0.76 ms.

The same set of performance metrics from the DNN model's case is provided for the BNN model predictions in Table \ref{tab:bnn_sinusoid}. Although the difference values are larger than those of the DNNs, this is likely due to the different RNGs used, despite the large volume of samples taken. Nonetheless, differences are small and do not indicate any fundamental issues within the backend mathematics.

\begin{table}[ht!]
    \centering
    \caption{Comparison of Fortran-based and TensorFlow-based BNN predictions on the noisy sinusoid problem.}
    \label{tab:bnn_sinusoid}
    \begin{tabular}{lccc}
        \toprule
        \textbf{Metric} & \textbf{Fortran} & \textbf{TensorFlow} & \textbf{Diff.} \\
        \midrule
        Mean AE                     & 1.215148   & 1.214302   & 0.000846  \\
        Max AE                      & 12.217185  & 12.298230  & $-0.081045$ \\
        Min AE                      & 0.000817   & 0.000489   & 0.000328  \\
        \midrule
        Mean APE (\%)               & 0.261509   & 0.261255   & 0.000254  \\
        Max APE (\%)                & 2.037004   & 2.049694   & $-0.012690$ \\
        % Min APE (\%)                & 0.000210   & 0.000167   & 0.000044  \\
        Std APE (\%)                & 0.294219   & 0.294237   & $-0.000018$ \\
        \midrule
        \textit{rRMSE} (\%)                  & 0.003935   & 0.003934   & 0.000002  \\
        $F_{\text{error}}>10\%$ (\%)            & 0.000000   & 0.000000   & 0.000000  \\
        $R^2$                       & 0.999591   & 0.999591   & 0.000000  \\
        \midrule
        \textbf{Total predictions}  & 1,000       & 1,000       & --        \\
        \textbf{Samples/pred.} & 20,000      & 20,000      & --        \\
        \textbf{CPU time (s)}       & 151.9441 & 1,215.8487 & -1{,}063.9046       \\
        \bottomrule
    \end{tabular}
\end{table}

\begin{figure}[ht!]
    \centering
    \includegraphics[width=\linewidth]{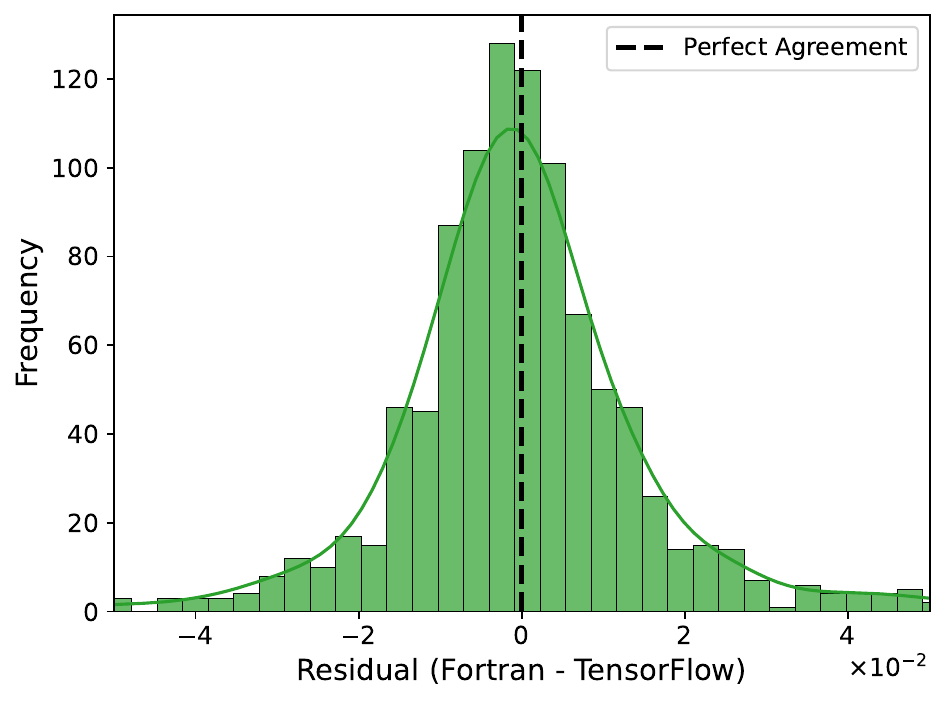}
    \caption{Histogram and KDE plot of the Fortran--Tensorflow BNN residual distribution.}
    \label{fig:bnn_histogram}
\end{figure}

The histogram and KDE plots for the residuals are provided in Fig.~\ref{fig:bnn_histogram}. Both are well centered at zero with the expected shape. The spread is relatively larger than that of Fig.~\ref{fig:dnn_histogram} but is still small compared with the magnitudes of the actual predicted values, which also range from 273.6 to 626.3.

\subsection{Test Case 2: Critical Heat Flux}

The second test considers CHF, the point in a boiling system at which there is a regime change from nucleate to transition boiling. CHF is a safety-related parameter in nuclear systems. This test was chosen as a final verification prior to the deployment of this framework within the CTF thermal-hydraulics code to support study of CHF surrogate modeling. To predict a CHF value, five inputs describing a tube channel are used: diameter, heated length, pressure, mass flux, and inlet subcooling. The dataset used for this experiment was a filtered version of the Nuclear Regulatory Commission CHF database used to construct the 2006 Groeneveld lookup tables~\cite{groeneveld2019critical} to only contain instances of dryout. This database was previously shown to yield satisfactory results when used to train a DNN~\cite{furlong2024hybrid}.

Both DNN and BNN models were trained in the TensorFlow environment, which achieved favorable error metrics on the test set: 2.18\% and 3.05\% relative errors with less than 3\% of all values greater than 10\% error. Both of these models are more complex than those used for the noisy sinusoid case. The DNN is structured with seven hidden layers, and the BNN is structured with four. Time per prediction is increased simply because of a larger number of calculations needed with increasing complexity. Once these models were confirmed to be well trained, they were exported and implemented with the Fortran framework to make predictions on the same test set (10\% of the filtered dataset). This implementation was configured identically to that which was later put into an experimental branch of CTF. As with the noisy sinusoid test case, the differences between the Fortran-based predictions and the TensorFlow-based predictions were taken in 9 key metrics, which are presented in Table~\ref{tab:chf_comparison}.

\begin{table}[ht!]
    \centering
    \caption{Fortran-based and TensorFlow-based implementation \textbf{difference values} on the CHF verification problem.}
    \label{tab:chf_comparison}
    \begin{tabular}{lcc}
        \toprule
        \textbf{Metric} & \textbf{DNN} & \textbf{BNN} \\
        \midrule
        Mean AE (\si{\kilo\watt\per\square\meter}) & 0.000011  & $-0.009558$ \\
        Max AE (\si{\kilo\watt\per\square\meter})  & $-0.000244$ & $-0.309853$ \\
        Min AE (\si{\kilo\watt\per\square\meter})  & $-0.000244$ & $-0.067495$ \\
        \midrule
        Mean APE (\%)                              & $-0.000001$ & $-0.000195$ \\
        Max APE (\%)                               & 0.000046  & $-0.011867$ \\
        % Min APE (\%)                               & 0.000000  & $-0.003226$ \\
        Std APE (\%)                               & 0.000002  & 0.004147  \\
        \midrule
        \textit{rRMSE} (\%)                        & 0.000000  & 0.000027  \\
        $F_{\text{error}}>10\%$ (\%)               & 0.000000  & 0.000000  \\
        $R^2$                                      & 0.000000  & $-0.000001$ \\
        \midrule
        \textbf{Total predictions}                 & 919       & 919 \\
        \textbf{Samples/pred.}                     & --         & 20{,}000 \\
        \textbf{Speedup factor}                    & 13.576 & 4.795 \\
        \bottomrule
    \end{tabular}
\end{table}

% Saving space by only presenting the diffs, rather than the actual values
% \begin{table}[ht!]
%     \centering
%     \caption{Comparison of Fortran-based and TensorFlow (TF)-based DNN predictions on the CHF prediction problem.}
%     \begin{tabular}{lccc}
%         \toprule
%         \textbf{Metric} & \textbf{Fortran} & \textbf{TF} & \textbf{Diff} \\
%         \midrule
%         Mean AE            & 32.932579  & 32.932568  & 0.000011  \\
%         Max AE             & 538.562500 & 538.562744 & -0.000244 \\
%         Min AE             & 0.068359   & 0.068604   & -0.000244 \\
%         \midrule
%         Mean APE (\%)      & 2.175051   & 2.175052   & -0.000001 \\
%         Max APE (\%)       & 33.239323  & 33.239277  & 0.000046  \\
%         Min APE (\%)       & 0.004131   & 0.004131   & 0.000000  \\
%         Std APE (\%)       & 2.400710   & 2.400708   & 0.000002  \\
%         \midrule
%         \textit{rRMSE} (\%)         & 0.032385   & 0.032385   & 0.000000  \\
%         Ferr > 10\% (\%)   & 1.088139   & 1.088139   & 0.000000  \\
%         $R^2$              & 0.996672   & 0.996672   & 0.000000  \\
%         \midrule
%         \textbf{Total Predictions} & 919 & 919 & -- \\
%         \textbf{CPU Time (s)} & 0.012224  & 0.165955 & -0.153731 \\
%         \bottomrule
%     \end{tabular}
% \end{table}

%%%%%%%%%%%%%%%%%%%%%%%%%%%%%%%%%%%%%%%%%%%%%%%%%%%%%%%%%%%%%%%%%%%%%%%%%%%%%%%%
\section{Conclusions}

This study presents a framework for implementing DNN and BNN models in Fortran, allowing for native execution without TensorFlow's C API, Python runtime, or ONNX conversion. Designed for ease of use and computational efficiency, the framework can be implemented in any Fortran code, supporting iterative solvers and UQ via ensembles or BNNs.

Verification was performed using a two-input, one-output test case composed of a noisy sinusoid to compare Fortran-based predictions to those from TensorFlow. The DNN predictions showed negligible differences and achieved a $19.6\times$ speedup, whereas the BNN predictions were observed with minor disagreement, plausibly because of differences in random number generation. An $8.0\times$ speedup was noted for BNN inference. The approach was then further verified on a nuclear-relevant problem predicting CHF, which demonstrated similar behavior along with significant computational gains. Discussion regarding the framework’s successful integration into the CTF thermal-hydraulics code is also included, outlining its practical usefulness. 

Overall, this framework was shown to be effective at implementing both DNN and BNN model inference within Fortran, allowing for the continued study of ML-based methods in real-world nuclear applications. Future work will include expanding support for Gaussian processes, enhancing UQ, and improving automation and compatibility with additional model formats.

%%%%%%%%%%%%%%%%%%%%%%%%%%%%%%%%%%%%%%%%%%%%%%%%%%%%%%%%%%%%%%%%%%%%%%%%%%%%%%%%
\section{Acknowledgments}

\noindent The authors from North Carolina State University were also funded by DOE's Office of Nuclear Energy Distinguished Early Career Program (DECP) under award number DE-NE0009467.

%%%%%%%%%%%%%%%%%%%%%%%%%%%%%%%%%%%%%%%%%%%%%%%%%%%%%%%%%%%%%%%%%%%%%%%%%%%%%%%%
\bibliographystyle{ans}
\bibliography{bibliography}
\end{document}